\documentclass[runningheads, envcountsame, a4paper]{llncs}
\usepackage{amsmath}
\usepackage{subfigure}
\usepackage{graphicx}			
\usepackage{epsfig} 			
\usepackage{mathptmx} 			
\usepackage{amsfonts}
\usepackage{amssymb}
\usepackage{epstopdf}
\usepackage[english]{babel}
\usepackage{pgfplots}
\usepackage{url}
\usepackage{lipsum}

\begin{document}
\title{Using Nonlinear Normal Modes for Execution of\\ Efficient Cyclic Motions in Articulated Soft Robots}
%
\author{Cosimo Della Santina\inst{1}, Dominic Lakatos\inst{1}, \\ Antonio Bicchi\inst{2}, Alin Albu\--Schaeffer\inst{1}%
}
\authorrunning{Della Santina et al.}
%
\institute{Institute of Robotics and Mechatronics, German Aerospace Center (DLR), Oberpfaffenhofen 82234, Germany, and Technical University of Munich \email{cosimodellasantina@gmail.com, alin.Albu-Schaeffer@dlr.de}
	\and
	``En\-ri\-co Pi\-ag\-gio'', U\-ni\-ver\-si\-ty of Pisa, Largo Lucio Lazzarino 1, 56126 Pisa, Italy, and Department of Advanced Robotics, Istituto Italiano di Tecnologia, via Morego, 30, 16163 Genova, Italy \\
}
\maketitle              

\begin{center}
	What follows is an \textbf{\color{red} old version} of the manuscript. You can find the \textbf{\color{blue} final one }at the following link:  \url{https://www.dropbox.com/s/x9a09gf38icu8sf/Using_nonlinear_normal_modes_ISER.pdf?dl=0}
	
	Feel free to get in touch via \texttt{c.dellasantina@tudelft.nl} if you cannot get access.
\end{center}

%
%
%
\section{Motivation, Problem Statement, Related Work} 

Inspired by the vertebrate branch of the animal kingdom, articulated soft robots are robotic systems embedding elastic elements into a classic rigid (skeleton\--like) structure \cite{della2020soft}.
Leveraging on their bodies elasticity, soft robots promise to push their limits far beyond the barriers that affect their rigid counterparts.
However, existing control strategies aiming at achieving this goal are either tailored on specific examples \cite{hutter2010slip}, or rely on model cancellations - thus defeating the purpose of introducing elasticity in the first place \cite{buondonno2015recursive,della2017controlling}.
%

In a series of recent works, we proposed to implement efficient oscillatory motions in robots subject to a potential field, aimed at solving these issues.
A main component of this theory are Eigenmanifolds, that we defined in \cite{albu2020tutorial} as nonlinear continuations of the classic linear eigenspaces.
When the soft robot is initialized on one of these manifolds, it evolves autonomously while presenting regular - and thus practically useful - evolutions, called normal modes.
In addition to that, we proposed in \cite{della2020control} a control strategy making modal manifolds attractors for the system, and acting on the total energy of the soft robot to move from a modal evolution to the other.
In this way, a large class of autonomous behaviors can be excited, which are direct expression of the embodied intelligence of the soft robot.

\begin{figure}[t]
\centering
\includegraphics[trim = {2cm 0 3.5cm 0}, clip, height = 0.435\columnwidth]{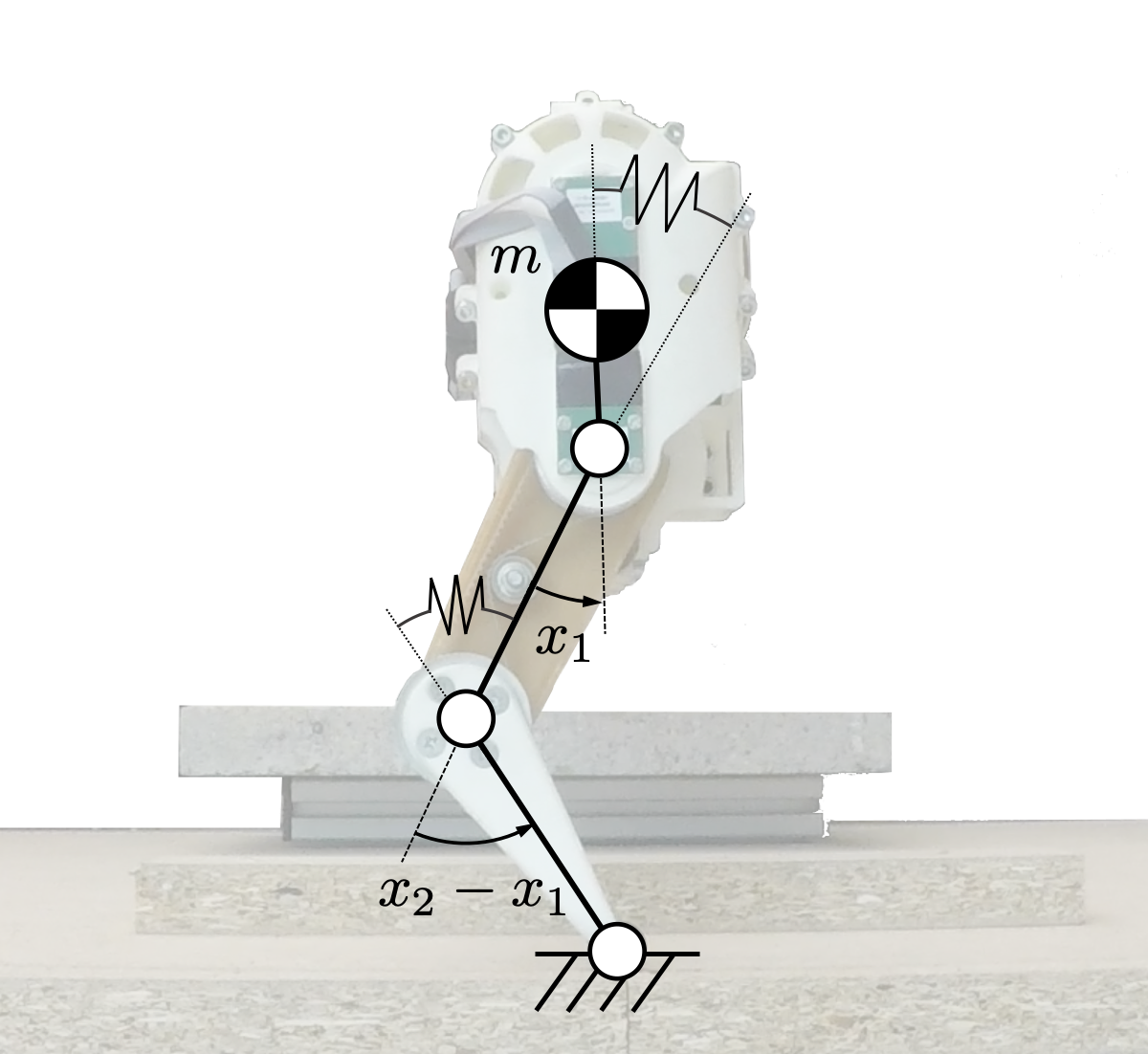}\hspace{0.0125\columnwidth}
\scalebox{-1}[1]{ \includegraphics[trim = {0 5cm 5cm 4cm}, clip, height = 0.435\columnwidth]{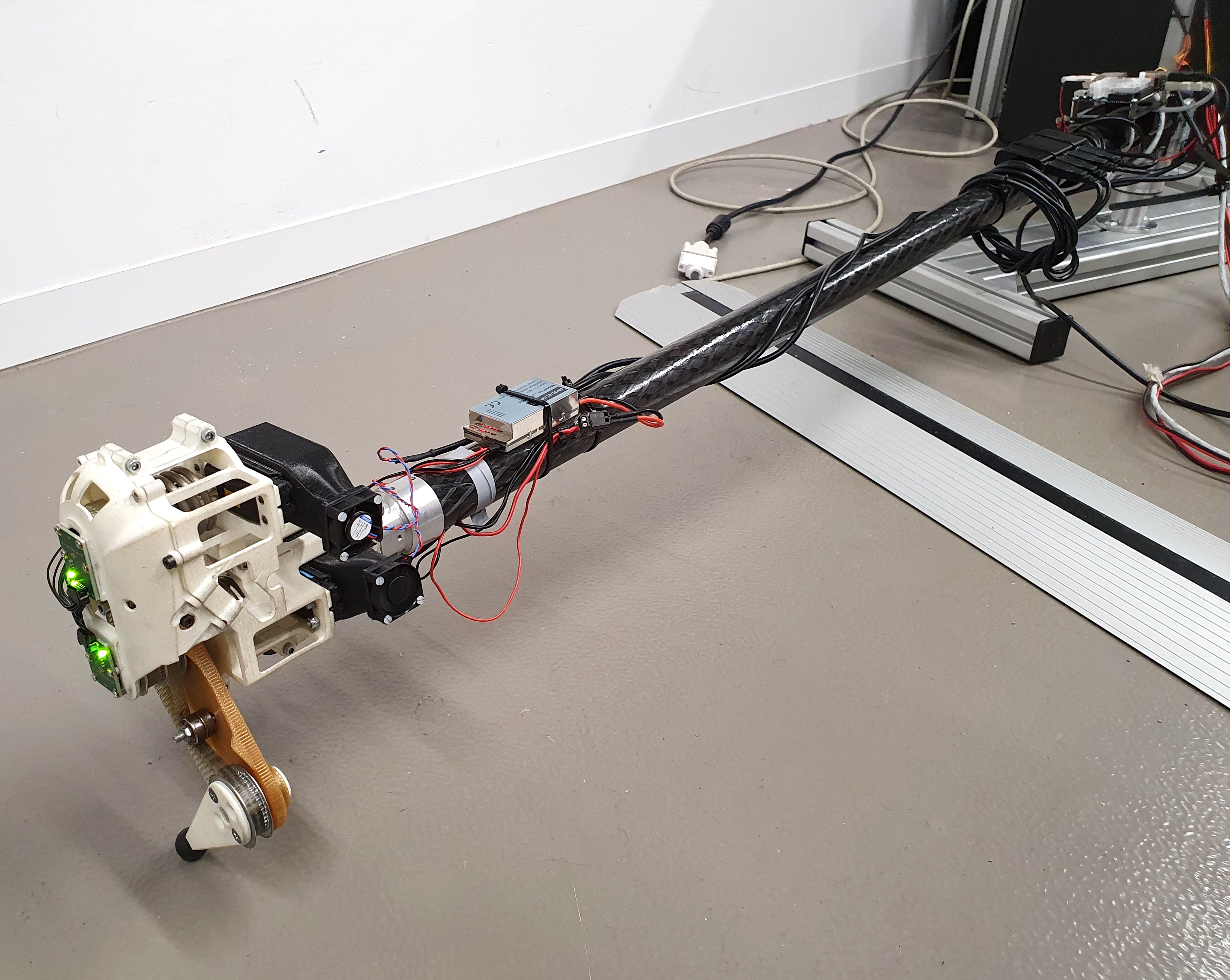} }
\caption{Experimental setup: a 2\--DoF (the upper part is constrained to stay vertical) segmented leg. The left panel shows also a sketch of the robot scheme with main quantities highlighted. The right panel shows the pole system constraining the upper part of the leg. \label{fig:segmented}}
\end{figure}

Despite the fact that  the idea behind our work comes from physical intuition and preliminary experimental validations \cite{lakatos2017eigenmodes}, the formulation that we have provided so far is however rather theoretical, and very much in need of an experimental validation.
The aim of this paper is to provide such an experimental validation using as testbed the articulated soft leg in Fig. \ref{fig:segmented}. We will introduce a simplified control strategy, and we will test its effectiveness on this system, to implement swing\--like oscillations.
We plan to extend this validation with a soft quadruped.

\section{Technical Approach}
\subsection{Eigenmanifold: a very concise definition}
Building upon a theory laid down by more than one century of research in mathematics, physics, and engineering, in \cite{albu2020tutorial} we propose an extension of linear modes to robotics, which is then summarized in a coordinate dependent framework in \cite{della2020control}.
We must give those definitions for granted here, due to space limitations, and only provide an intuitive introduction to the concept.
Consider a mechanical system in the standard form
%
$M(x)\ddot{x} + C(x,\dot{x})\dot{x} + G(x) = \tau$,
%
where $x \in \mathbb{R}^n$ are the configuration coordinates of the robot. 
$M(x),C(x,\dot{x}) \in \mathbb{R}^{n \times n}$ are the usual inertia and  Coriolis matrices, and $G(x)  \in \mathbb{R}^{n}$ is the potential field. $\tau \in \mathbb{R}^n$ are the control inputs. The total energy is $E(x,\dot{x}) = x^{\mathrm{T}}M(x)x/2 + V(x)$, where $V(x)$ is the potential associated to $G(x)$.

An Eigenmanifold is a direct extension of an Eigenspace to this kind of nonlinear mechanical systems. It is defined by imposing to a curved surface many of the properties that define an eigenspace in the linear case.
It is a two dimensional invariant submanifold of the configuration space $(x,\dot{x})$, which contains an equilibrium configuration of the robot. Also, it is such that any evolution contained in it is periodic, it has a trajectory which is line\--shaped, and it is unequivocally identified by its energy level. 
Consider an eigenspace of the linearized system at an equilibrium $ES = \mathrm{Span}\left\{(c,0),(0,c)\right\}$, with $c\in\mathbb{R}^n$. In \cite{albu2020tutorial} we show that we can always describe the Eigenmanifold prolonging this linear eigenspace as the set of states such that
\begin{equation} \small
	X(x_{\mathrm{m}},\dot{x}_{\mathrm{m}}) = x, \quad \dot{X}(x_{\mathrm{m}},\dot{x}_{\mathrm{m}}) = \dot{x},
\end{equation}
where $X$ and $\dot{X}$ are two functions from $ES$ to $\mathbb{R}^n$ describing the manifold geometry - called coordinate embedding - and $(x_{\mathrm{m}},\dot{x}_{\mathrm{m}}) = (c^{\mathrm{T}}x, c^{\mathrm{T}}\dot{x})$.
In \cite{albu2020tutorial} we discuss how extracting $(X,\dot{X})$ from $ES$.


\subsection{Exciting nonlinear oscillations with a simple feedback}

Let $(X,\dot{X})$ be a coordinate embedding of an eigenmanifold. In \cite{della2020control} we proposed the following feedback control to excite nonlinear oscillations in a robot subject to a potential field
\begin{equation}\label{eq:fb_stabilization}
\small
\tau(x, \dot{x}) = 
M(x)\left( \kappa_\mathrm{p} \left(X(x_{\mathrm{m}}, \dot{x}_{\mathrm{m}}) - x\right) + \kappa_\mathrm{d} \, \left(\dot{X}(x_{\mathrm{m}}, \dot{x}_{\mathrm{m}}) - \dot{x}\right) + \alpha\tau_{\mathrm{E}}(x,\dot{x}, \bar{E}) \right).
\end{equation}
The idea is to make the modal manifold an attractor by means of a PD\--like action (first two terms), and then pick the right oscillation among all the available ones through energy regulation (implemented by $\tau_{\mathrm{E}}$).
The control gains are $\kappa_\mathrm{p},\kappa_\mathrm{d},\alpha \in \mathbb{R}$. 
The PD regulation is quite simple to implement, since $X$ and $\dot{X}$ are just polynomial functions. Due to this simplicity, we experienced an high level of robustness when testing it in simulation. We want to double\--check this intuition experimentally here.

\begin{figure}[t]
	\centering
		\includegraphics[height = .4\columnwidth]{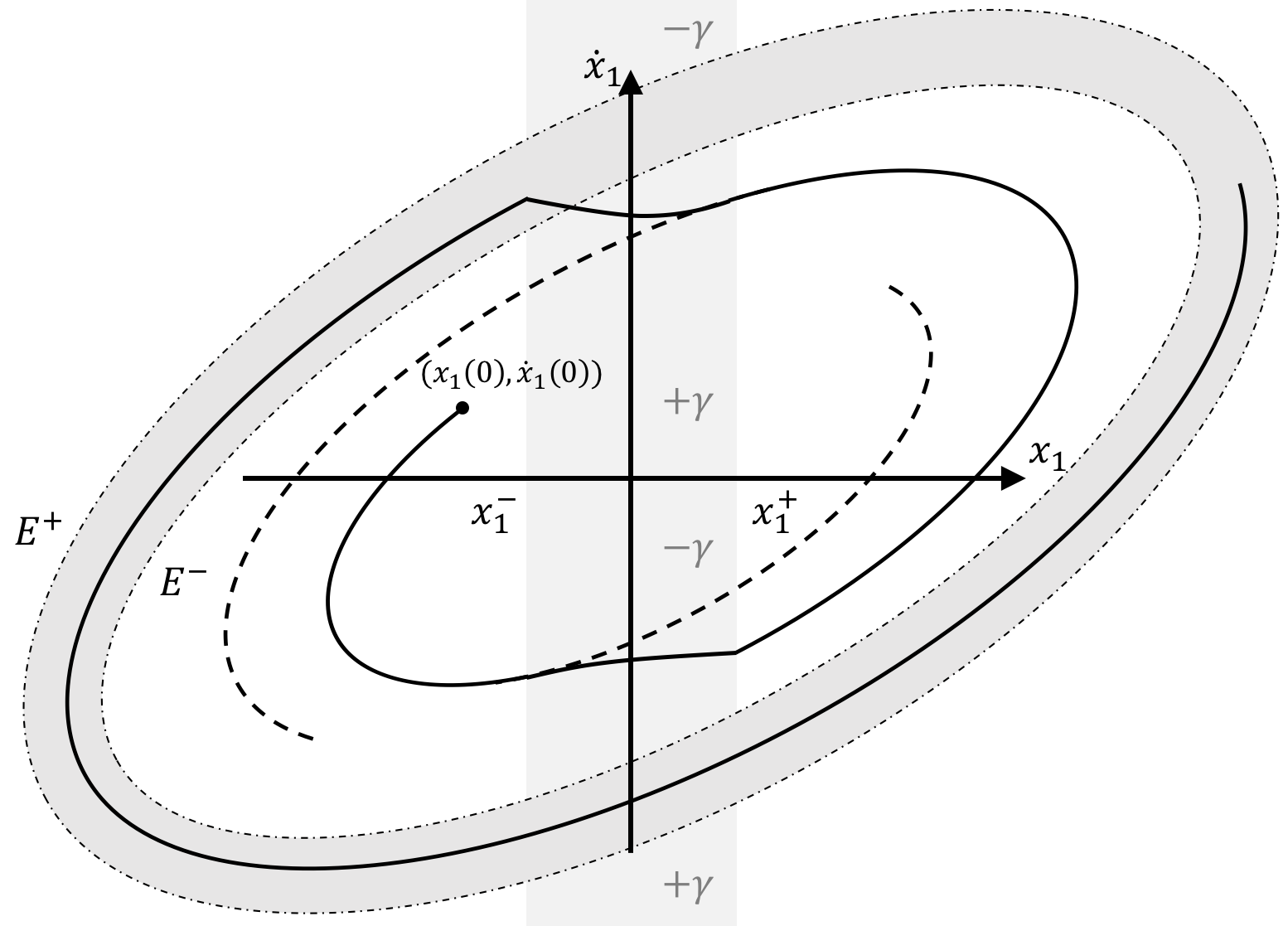}
	\caption{The state evolves under the control action \eqref{eq:bar_tau}. 
		When the system is in a neighborhood of equilibrium configuration $x_{\mathrm{m}} = 0$ (i.e. when it crosses the gray area), energy is injected by the controller, moving the system to another of its autonomous orbits. Eventually this brings the robot in the region of state space with the desired amount of energy. \label{fig:bangbang}}
\end{figure}

The possibility of injecting or removing energy from the system allows to select the desired mode within the modal manifold, therefore increasing or decreasing the amplitude of oscillation.
In \cite{della2020control}, $\tau_{\mathrm{E}}(x,\dot{x}, \bar{E}) $ realizes energy regulation through the feedback loop $\dot{x} \left(\bar{E} - E(x,\dot{x})\right)$.
%
%
This is a more complex component to implement since the energy is a transcendental function of the state, and fundamentally realizes a form of strongly model dependent feedback.
Since we want to bring this abstract theory on an experimental ground, we consider here an even simpler and more robust strategy. This is effectively the energy regulation introduced in \cite{della2020control}, reduced to its most essential components
%
%
		\begin{equation}\label{eq:bar_tau}
		\small
		\tau_{\mathrm{E}} \! = \!
		\begin{cases}
		\;\;\; 0 \;\; \mathrm{if} \; x_{\mathrm{m}} \notin [x^-_{\mathrm{m}}, x_{\mathrm{m}}^+] \; \vee \; E(X,\dot{X}) \in [E^- , E^+] \\
		\;\;\; 1 \;\; \mathrm{if} \;  x_{\mathrm{m}} \in [x^-_{\mathrm{m}}, x_{\mathrm{m}}^+] \; \wedge  \; ((E(X,\dot{X}) < E^- \, \wedge \, \dot{x}_{\mathrm{m}} > 0) \vee (E(X,\dot{X})> E^+ \, \wedge \, \dot{x}_{\mathrm{m}} < 0)) \\
		-1 \;\; \mathrm{otherwise}
		\end{cases},
		\end{equation}
		where $E^+ >E^- > 0$, and $x_{\mathrm{m}}^+ > 0 > x_{\mathrm{m}}^-$ are scalar constants. 
		This controller idea is shown in Fig. \ref{fig:bangbang}. Similar to a swing which is kept in persistent oscillations by an occasion push in the right direction, the idea is to inject or remove energy in small chunks until $E(x(t))$ reaches $[E^-, E^+]$. 
		Note indeed that the conditions selecting the sign of the torque are such that the change of energy $\dot{E} = \dot{x}^T \tau_{\mathrm{E}}$ is always positive when $E(X,\dot{X}) < E^-$ and negative viceversa. Note that this strategy is akin to the well\--known swing up controller proposed in  \cite{aastrom2000swinging}.
		
		Although built with the goal of being intuitive and robust, this control action makes the closed loop system hybrid. Therefore the actual proof of convergence will require some effort which is beyond the scope of the present paper. Our aim here is instead to give experimental substantiation to the whole idea of exciting complex nonlinear oscillations by means of simple feedback control actions stabilizing Eigenmanifolds, and see which kind of lessons we can learn from this validation.

\section{Experimental validation} 

\subsection{Experimental setup}

As a first experimental validation of the proposed strategy, we consider here the soft segmented leg in Fig. \ref{fig:segmented}.
It is made of two links with same length $b$ - considered here massless - and a main body - with mass $m$. 
Linear revolute springs act on both joints (mechanism not shown in figure).
Please refer to \cite{lakatos2017eigenmodes} for more details on this system.
The leg is mechanically constrained to evolve on the Sagittal plane and the main body to remain vertical, by the pole shown in the right part of Fig. \ref{fig:segmented}. We hypothesize infinite friction between the foot and the environment, so that the ground contact behavior is approximated with a revolute joint. 

\subsection{The model, the mode, and the control algorithm}

We are interested here in generating swing oscillations. 
%
We describe the system configuration through polar coordinates of the center of mass expressed w.r.t. the foot frame
%
		$\theta = x_1/2 + x_2/2$, 
		$r = b \, \sqrt{2 \, (1 + \cos{(x_1-x_2)})}$.
%
The resulting dynamics has the following form 
%
		$\ddot{\theta} = -2\dot{r}\dot{\theta}/r + g\sin(\theta)/r - 2 \gamma \theta/r^2 + \tau_{\mathrm{\theta}}/m$, and
		$\ddot{r} = r \dot{\theta}^2- g \cos(\theta) - \gamma (\Upsilon(r) - \Upsilon(r_0)) /\sqrt{4 b^2 - r^2} + \tau_{\mathrm{r}}/m$
%
where $\Upsilon(r) = \arccos\left(1 - r^2/(2 b^2)\right)$, $\gamma$ is the stiffness of both springs divided by the mass $m$, $\tau_\mathrm{\theta}$ and $\tau_\mathrm{r}$ are the control actions, $g$ is the gravity acceleartion, and $r_0$ is the unloaded length of the equivalent spring. 
%
%
The system has an equilibrium in $\theta = 0$ and $\Upsilon(r) = \Upsilon(r_0) - g/\gamma$. Its linearized dynamics is $\Delta\ddot{\theta} \simeq k_{\theta} \Delta\theta$, and $\Delta \ddot{r} \simeq k_{\mathrm{r}} \Delta r$, with $k_{\theta}, k_{\mathrm{r}} \in \mathbb{R}$ being two constant values. The normal modes of the linearized system are therefore two decoupled evolutions: an angular oscillation with fixed radius, and a radial oscillation with fixed angle.
The nonlinear extension of the latter is trivial, since for $\theta \equiv 0$ and $\dot{\theta} \equiv 0$ the dynamics collapses into a quasi\--linear one, that we studied in \cite{lakatos2017eigenmodes}.
The other mode instead turns into a more complex oscillation, that we investigate here.
In this case $c = (1,0)$, i.e. $x_{\mathrm{m}} = \theta$.
We approximate $(X,\dot{X})$ as fourth order polynomials, solving in the Galerkin sense the tangency constraints introduced in \cite{della2020control}.

\subsection{Experiments completed}

\begin{figure*}[t]
	\centering
	\includegraphics[width=0.9\columnwidth]{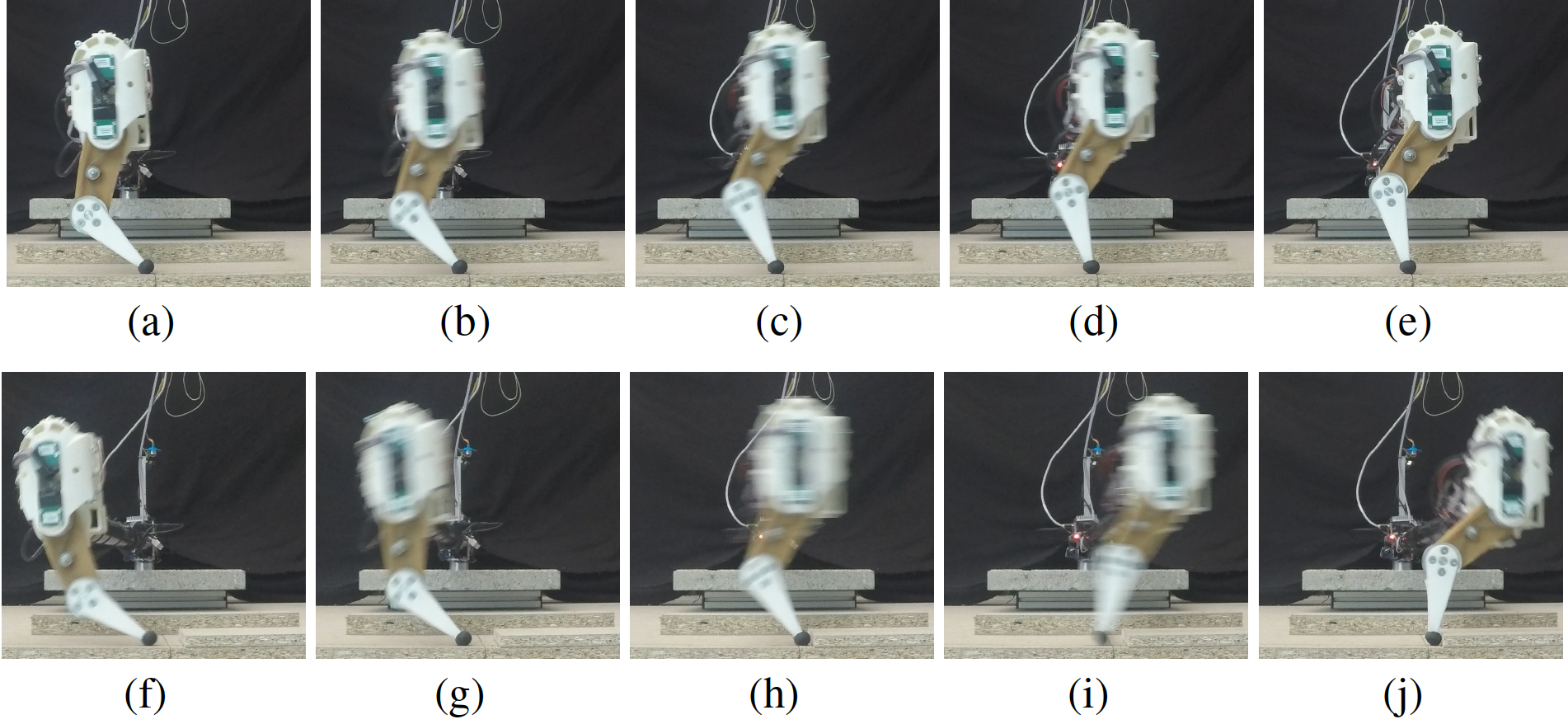}
	\vspace{-0.5cm}
	\caption{{Nonlinear oscillations induced by the proposed algorithm on a segmented soft leg. Panels (a-e) present one oscillation for $\alpha = 0.3$Nm, while panels (f-j) show the case of $\alpha = 0.9$Nm.}
		{\color{red} Quality of the figure reduced for the ArXiv version} \label{fig:photoseq_leg}}
\end{figure*}

\begin{figure*}[t]
	\centering
	\includegraphics[trim = {0 0 0 0}, clip, width = .8\columnwidth]{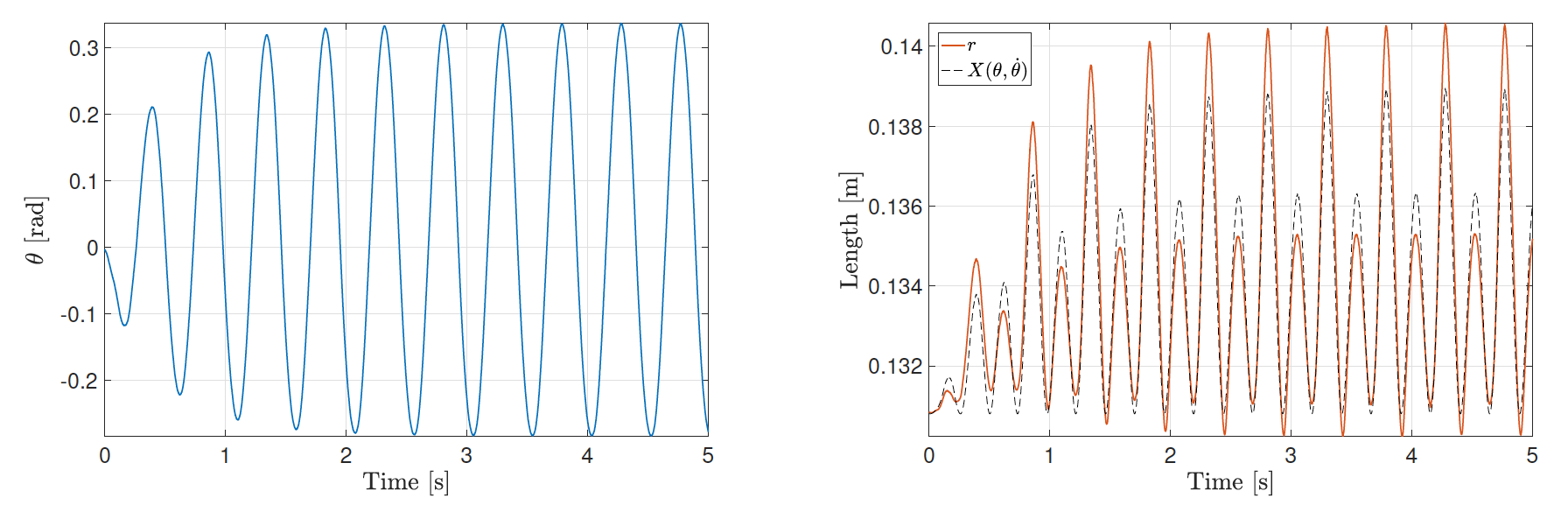}
	\vspace{-0.3cm}
	\caption{{Experimental evolutions of $(\theta,r)$, for $\alpha = 0.5\text{Nm}$. The right panel reports also the ideal evolution on the manifold $X(\theta,\dot{\theta})$ for the measured evolution of $\theta$, as a dashed gray line. {\color{red} Quality of the figure reduced for the ArXiv version}
		} \label{fig:segmented_polar}}
\end{figure*}

\begin{figure}[t]
	\centering
\includegraphics[width = .85\columnwidth]{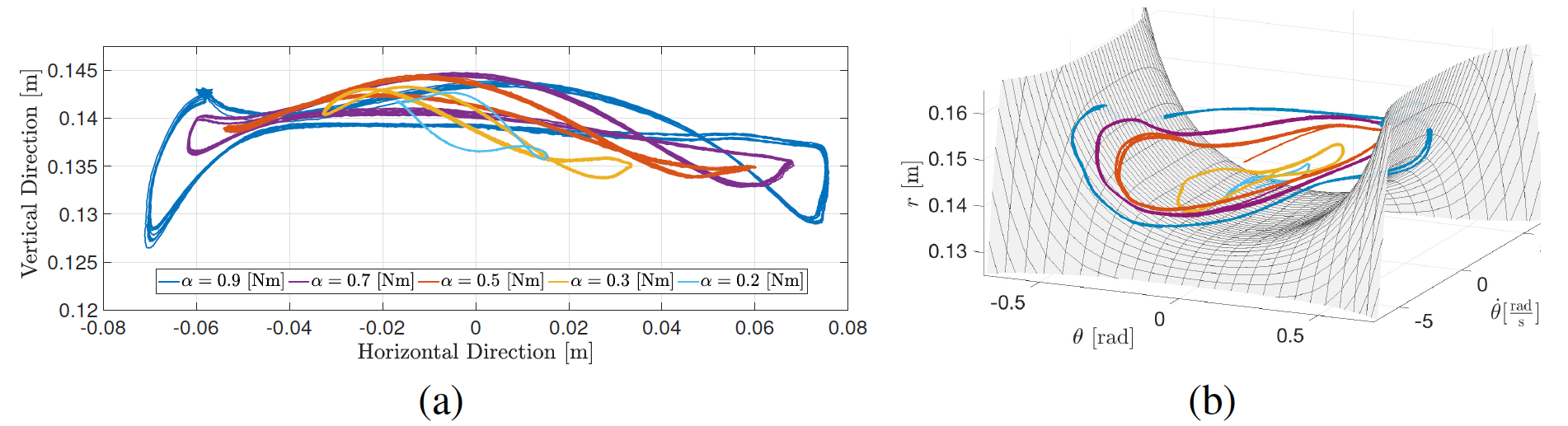} 
	\vspace{-0.5cm}
	\caption{Experimental trajectories 
	Panel (a) shows the evolutions in Cartesian coordinates of the leg's center of mass. Panel (b) presents the same evolutions in the space $(\theta,\dot{\theta},r)$. $15$s of oscillations are considered. The ideal modal manifold $r = X(\theta,\dot{\theta})$ is superimposed. 
	{\color{red} Quality of the figure reduced for the ArXiv version}
	\label{fig:segmented_cartesian}
	\label{fig:manifold_ev_ex}
} 
\end{figure}

We performed experiments for five different values of the orbit excitation gain $\alpha$; $0.2$Nm, $0.3$Nm, $0.5$Nm, $0.7$Nm, $0.9$Nm. Target energy levels are  $E^- = 21 \text{J}$, $E^+ = 22 \text{J}$.  Due to dissipation, the desired level of energy could not be reached. Instead energy injected through $\tau_{\mathrm{E}}$ is compensated by dissipation. A different equilibrium is reached for each value of $\alpha$.
Fig.~\ref{fig:photoseq_leg} shows oscillations resulting from two of the considered gains.
Fig.~\ref{fig:segmented_polar} shows the evolution of $\theta$ and $r$ for $\alpha = 0.5\text{Nm}$. 
Control action is turned on at $0$s. After a short transient lasting for about $2$s, in which the algorithm pumps energy into the system, the segmented leg starts to evolve according to a stable nonlinear oscillation.
Actual and ideal  trajectories - i.e. $(\theta,\dot{\theta},r,\dot{r})$ and $(\theta,\dot{\theta},X(\theta,\dot{\theta}),\dot{X}(\theta,\dot{\theta}))$ respectively - are quite close to each other, as shown by the right panel.
Fig. \ref{fig:segmented_cartesian} shows the evolutions of the center of mass in Cartesian coordinates, for all the considered values of $\alpha$, and for a period of $15$s.  The bigger is the gain, the larger are the oscillations, and the higher is the energy level reached. The resulting oscillations 
and highly repeatable.
Fig. \ref{fig:manifold_ev_ex} illustrates the evolutions superimposed to the ideal modal manifold, i.e. to the surface $(\theta,r) = X(\theta, \dot{\theta})$.
%

\subsection{Experiments scheduled}

We plan to test the behavior of the system under external disturbances, to see if the algorithm is robust even to these conditions.
Finally, we plan to test the algorithm on a quadruped built using four of the above discussed soft segmented legs, to see if the proposed control strategy can excite stable oscillations also in this more complex system. 
We will also consider of preliminarily investigating if such oscillations can be used to implement locomotion patterns.

%

\section{Experimental Insights} 

The experiments exhibit a quite different scenario than the one we could have expected by looking at the problem from the pure lenses of theory.
For a start, they suggest that the proposed strategy can be used in practice to excite the normal modes of soft robots, generating stable and repeatable nonlinear oscillations also in the presence of many uncertainties and unmodeled dynamics in the controlled system - e.g. the actuators dynamics, the moving contact with the ground,  the non zero weight of the legs, inexact identification of system parameters, neglected friction effects. 
Moreover, $\dot{x}_1,\dot{x}_1$ is not being directly measured, but estimated through a high pass filter. Finally, the physical system is serially actuated, and the parallel elastic behavior needs to be implemented through an opportune input mapping. First, we map $\tau_{\mathrm{\theta}}$ and $\tau_{\mathrm{r}}$ to torques acting on $x_1$ and $x_2$ via pre\--multiplication for the transpose Jacobian of the change of variables. These torques are then realized by commanding to motors a displacement equal to the torques divided by the stiffness $\gamma$.
%
%
%
These experiments also taught us some very important lessons on where to look for improvements. First and foremost, the final energy resulted as an equilibrium between $\alpha$ and dissipative effects, rather than due to some stopping condition connected to the energy level $E^{-}$ that could never be reached. This is because the energy regulator was built with a conservative system in mind. 
So experiments suggest that rather than improving the Eigenmanifold (e.g. to make it more global), it is probably more practically meaningful to put some effort in developing adaptive algorithms which can dynamically adjust the value of $\alpha$ so to reach a desired amplitude of oscillations.
For what concerns the specific leg experiment instead, we saw that mismatches from the manifold start to increase with high velocities, and that the oscillations are typically not perfectly symmetric (higher errors for positive values of $\delta$). This suggests that a model taking into account the legs mass should be considered, therefore breaking the symmetry of $(X,\dot{X})$.
Finally, we expect the scheduled experiments to provide further insights in how to make the controller even more robust, and on how to scale it up to very high dimensional conditions.

\bibliographystyle{splncs04}
\bibliography{biblio.bib,biblio_2.bib}

\end{document}